\documentclass[conference,a4paper]{IEEEtran}
\IEEEoverridecommandlockouts
\usepackage[utf8]{inputenc}
\usepackage[T1]{fontenc}
\usepackage[english,vietnamese]{babel}

\usepackage{ifpdf}
\ifCLASSINFOpdf
  \usepackage[pdftex]{graphicx}
  \graphicspath{{../pdf/}{../jpeg/}}
  \DeclareGraphicsExtensions{.pdf,.jpeg,.png,.bmp}
\else
  \usepackage[dvips]{graphicx}
  \graphicspath{{../jpg/}}
  \DeclareGraphicsExtensions{.jpg}
  \fi
\usepackage[cmex10]{amsmath}
\usepackage{array}
\usepackage{mdwmath}
\usepackage{mdwtab}
\usepackage{mathrsfs}
\usepackage{amssymb}
\usepackage{eqparbox}
\usepackage{epstopdf}
\usepackage{url}
\usepackage{cite}
\usepackage{subfiles}
\usepackage{soul,color}
\usepackage{amsthm}
\usepackage{hyperref}
\usepackage{balance}
\usepackage{cases} 
\makeatletter
\def\ScaleIfNeeded{\ifdim\Gin@nat@width>\linewidth\linewidth\else\Gin@nat@width\fi}

\usepackage{pifont}
\usepackage{xcolor}

\usepackage{graphicx}        

\usepackage{subcaption}

\usepackage{algorithm}
\usepackage{algpseudocode}

\usepackage{booktabs}
\usepackage{multirow}
\usepackage{multicol}

\hypersetup{
    colorlinks   = true,
    citecolor    = blue,
    linkcolor    = blue,
    urlcolor     = black
}

\allowdisplaybreaks

\begin{document}
\selectlanguage{english}
\setlength{\belowdisplayskip}{5pt} \setlength{\belowdisplayshortskip}{5pt}
\setlength{\abovedisplayskip}{5pt} \setlength{\abovedisplayshortskip}{5pt}

\title{{Less Is More: A Leakage-Controlled Study of Dermoscopic Preprocessing for Joint Skin Lesion Classification and Segmentation with YOLO26}
\thanks{Corresponding author: Nguyen Thai Anh (email: anh.nt@vlu.edu.vn).}
}

\author{
\IEEEauthorblockN{\resizebox{.99\linewidth}{!}{Truong Viet Vu, Nguyen Chi Hai, Nguyen Phuc Nguyen, Ngo Hoang Tu, Vo Nguyen Quoc Bao, and Nguyen Thai Anh}}
\IEEEauthorblockA{
Faculty of Information Technology, Van Lang School of Technology, Van Lang University, Ho Chi Minh City, Vietnam\\
Email: vu.2274802011045@vanlanguni.vn, hai.2274802010214@vanlanguni.vn, nguyen.2274802010586@vanlanguni.vn,\\ tu.nh@vlu.edu.vn, bao.vnq@vlu.edu.vn, anh.nt@vlu.edu.vn}
}

\maketitle

\begin{abstract}
Handcrafted preprocessing is widely employed in automated dermoscopic analysis to suppress imaging artifacts and enhance lesion visibility. Nevertheless, its actual contribution to modern real-time models remains unclear, particularly when evaluation protocols do not adequately control correlations among images of the same lesion. This study presents a leakage-controlled, lesion-disjoint evaluation of dermoscopic preprocessing and augmentation for joint multi-class lesion classification and instance segmentation using a fixed nano-scale YOLO26 segmentation model (YOLO26n-seg). From HAM10000 (10,015 images), quality control yields 10,013 valid image--mask pairs from 7,468 unique lesions, partitioned into mutually exclusive sets by lesion identity. With the architecture, resolution, training budget, and evaluation protocol held fixed, we compare minimally processed images plus online augmentation against offline class balancing, DullRazor--CLAHE preprocessing, and raw--processed hybrid views, over three random seeds. On the lesion-disjoint test set, the raw baseline achieves a mask mAP$_{50:95}$ of $0.5636 \pm 0.0234$, a Dice score of $0.9356 \pm 0.0024$, and a macro-F1 score of $0.6917 \pm 0.0202$. Offline augmentation does not improve the mean performance, while the combined and hybrid strategies reduce both class-aware segmentation and classification accuracy. At only 2.69 million parameters, the model runs at approximately 50 frames per second. Under a leakage-controlled, lesion-disjoint protocol with all non-input factors held fixed, minimally processed dermoscopic images combined with standard online augmentation deliver a better accuracy--efficiency trade-off than increasingly complex deterministic preprocessing, which yields no consistent joint benefit across three seeds on HAM10000.
\end{abstract}

%

\vspace{-0.15cm}
\section{Introduction}

Skin lesion assessment relies strongly on visual characteristics such as color, texture, shape, and border structure. Dermoscopy provides non-invasive visualization of subsurface features and improves melanoma diagnostic accuracy over unaided examination \cite{kittler2002diagnostic}, yet interpretation remains challenging because different lesion categories may appear similar while lesions within a category vary considerably. Deep neural networks have therefore emerged as decision-support tools that learn discriminative representations directly from dermoscopic images \cite{esteva2017dermatologist}.

Automated dermoscopic analysis commonly involves lesion segmentation and classification. Segmentation delineates abnormal tissue, whereas classification assigns the lesion to a diagnostic category. Research on both tasks has been accelerated by public challenges organized by the International Skin Imaging Collaboration (ISIC) \cite{codella2019skin}. The Human Against Machine with 10,000 training images (HAM10000) dataset is a widely used benchmark comprising 10,015 dermoscopic images from seven diagnostic categories \cite{tschandl2018ham10000}. Because multiple images may represent the same lesion under different acquisition conditions, random image-level partitioning can place correlated observations in separate subsets and yield optimistic estimates of generalization. Lesion-level separation is therefore important for reliable medical machine-learning evaluation \cite{rouzrokh2022mitigating}.

Dermoscopic images may contain hairs, uneven illumination, low contrast, and acquisition artifacts. Common preprocessing techniques include DullRazor hair removal \cite{lee1997dullrazor}, contrast-limited adaptive histogram equalization (CLAHE) \cite{zuiderveld1994contrast}, Gray World color normalization \cite{buchsbaum1980spatial}, and lesion-centered region-of-interest (ROI) extraction. Although visually motivated, these transformations may alter diagnostically relevant information or introduce artificial patterns. Consistent with this concern, Mahbod \textit{et al.} found that segmentation-based image manipulation did not consistently improve dermoscopic classification and that removing surrounding context could reduce performance \cite{mahbod2020effects}.

Data augmentation raises a related concern. Online augmentation applies stochastic transformations during training, whereas offline augmentation explicitly creates additional samples that can target minority classes but remain correlated with their sources and must be restricted to the training set \cite{shorten2019survey}. Distinguishing the two is essential to avoid confounding preprocessing, sampling, and augmentation effects.

Classification and segmentation are often studied separately, although both objectives can be learned from shared dermoscopic representations \cite{yang2017novel}. Their performance is not interchangeable: accurate lesion overlap does not guarantee correct diagnosis, and correct classification may accompany an incomplete boundary. A joint system should therefore be assessed using both spatial and class-aware metrics \cite{nguyen2025swin,vu2026wood}. The YOLO family provides an efficient foundation for this setting by combining localization and class prediction within a unified framework \cite{redmon2016you}, and its recent extensions add instance segmentation while preserving practical inference speed. In particular, the nano-scale YOLO26 segmentation variant (YOLO26n-seg) jointly predicts lesion location, diagnostic class, confidence, and a pixel-level mask \cite{jocher2026ultralytics}. Comparisons among recent YOLO generations further demonstrate that architecture and model scale affect accuracy--latency trade-offs, emphasizing the need for matched experimental conditions \cite{pham2025fruit,hai2025hybrid}.

Despite these advances, the benefit of dermoscopic preprocessing remains unclear because previous evaluations often vary data partitioning, architecture, resolution, augmentation, sampling, and prediction policies simultaneously. It is therefore necessary to determine whether increasingly complex preprocessing provides a genuine advantage under leakage-controlled and otherwise matched conditions.

This study evaluates handcrafted dermoscopic preprocessing for lightweight joint multi-class lesion classification and instance segmentation under a lesion-disjoint protocol. Using a fixed YOLO26n-seg architecture, we compare minimally processed images, offline class-balancing augmentation, DullRazor--CLAHE preprocessing, individual preprocessing components, ROI transformations, and hybrid raw--processed training views, evaluating the principal configurations across three random seeds.

The main contributions of this study are as follows:

\begin{itemize}
    \item \textbf{Leakage-controlled evaluation:} We show that the pre-existing image-level split leaks 765 lesions (of 7,468) across subsets, and we construct a SHA-256--frozen, lesion-disjoint partition with zero cross-subset overlap.

    \item \textbf{Controlled preprocessing analysis:} Holding the architecture, resolution, training budget, thresholds, and evaluation procedure fixed, we isolate preprocessing and augmentation as the only varied factors across four principal configurations and seven component ablations.

    \item \textbf{Finding and reproducible baseline:} Over three seeds we jointly report classification, class-aware segmentation, boundary quality, model size, and speed, and we show that minimally processed inputs with online augmentation give the best accuracy--efficiency trade-off, establishing a reproducible lightweight baseline (2.69\,M parameters, approximately 50 FPS).
\end{itemize}

The remainder of this paper is organized as follows. Section II presents the dataset preparation, preprocessing configurations, training protocol, and evaluation criteria. Section III reports and discusses the experimental results. Section IV concludes the paper and outlines directions for future research.

\section{Methodology}

\subsection{Dataset and Annotations}

The experimental corpus is constructed by matching HAM10000 metadata, dermoscopic images, and segmentation masks through \texttt{image\_id}. HAM10000 contains 10,015 images from seven classes \cite{tschandl2018ham10000}: actinic keratoses and intraepithelial carcinoma (\textit{akiec}), basal cell carcinoma (\textit{bcc}), benign keratosis-like lesions (\textit{bkl}), dermatofibroma (\textit{df}), melanoma (\textit{mel}), melanocytic nevi (\textit{nv}), and vascular lesions (\textit{vasc}). Corresponding publicly released ISIC segmentation masks, which serve as the segmentation ground truth, are matched by image identifier \cite{codella2019skin}.

Quality control excludes two uniformly foreground masks, \texttt{ISIC\_0026042} and \texttt{ISIC\_0029819}, because they do not delineate lesion boundaries. The final corpus therefore contains 10,013 valid image--mask pairs from 7,468 lesions, distributed as shown in Table~\ref{tab:dataset_distribution}.

\begin{table}[t]
\caption{Class distribution of the lesion-disjoint dataset.}
\label{tab:dataset_distribution}
\centering
\small

\begingroup
\renewcommand{\arraystretch}{1.15}

\begin{tabular}{lrrrr}
\hline
Class & Train & Validation & Test & Total \\
\hline
\textit{akiec} & 229 & 49 & 49 & 327 \\
\textit{bcc}   & 360 & 77 & 77 & 514 \\
\textit{bkl}   & 768 & 165 & 165 & 1,098 \\
\textit{df}    & 80 & 18 & 17 & 115 \\
\textit{mel}   & 778 & 167 & 167 & 1,112 \\
\textit{nv}    & 4,693 & 1,006 & 1,006 & 6,705 \\
\textit{vasc}  & 98 & 22 & 22 & 142 \\
\hline
\textbf{Total images} &
\textbf{7,006} &
\textbf{1,504} &
\textbf{1,503} &
\textbf{10,013} \\
\hline
\end{tabular}

\endgroup
\end{table}

Each binary mask is converted into a class-aware YOLO polygon from its largest valid external contour and simplified using the Douglas--Peucker algorithm \cite{douglas1973algorithms}. The audit verifies image and mask readability, polygon validity, label completeness, and consistency with diagnostic metadata.

\subsection{Lesion-Disjoint Split and Leakage Audit}

HAM10000 contains multiple images of some lesions; consequently, image-level splitting can place correlated observations in different subsets and overestimate generalization \cite{rouzrokh2022mitigating}. Indeed, the pre-existing image-level split contains 765 lesions spanning more than one subset.

We therefore group samples by \texttt{lesion\_id} and generate an approximately class-stratified 70\%/15\%/15\% split using seed 42. The training, validation, and test sets contain 7,006, 1,504, and 1,503 images from 5,231, 1,117, and 1,120 lesions, respectively. All classes are represented in each subset, with zero lesion overlap. The assignments and file provenance are frozen in a manifest whose split-defining contents are recorded using a 256-bit Secure Hash Algorithm (SHA-256) digest.

\begin{figure}[t]
    \centering
    \includegraphics[width=0.85\columnwidth]{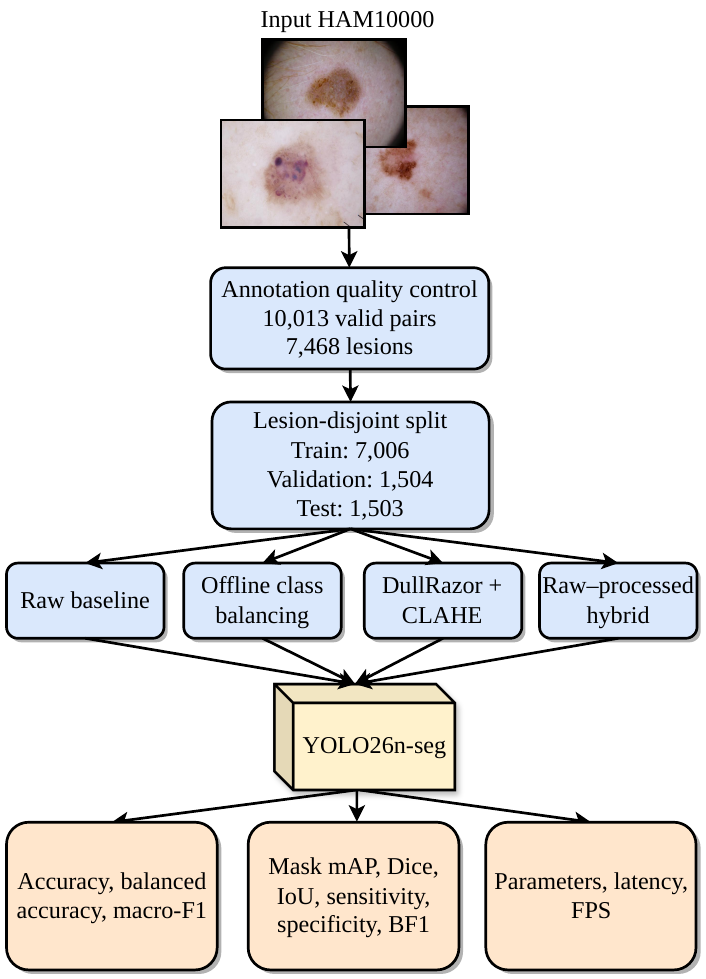}
    \caption{Overview of the experimental pipeline.}
    \label{fig:methodology_pipeline}
\end{figure}

\subsection{Preprocessing and Augmentation Strategies}

Fig.~\ref{fig:methodology_pipeline} summarizes the pipeline, and Table~\ref{tab:input_configurations} defines the four principal input strategies. Images and masks are letterboxed to $640\times640$ pixels using bilinear and nearest-neighbor interpolation, respectively.

\begin{table*}[t]
\caption{Principal preprocessing and augmentation configurations.}
\label{tab:input_configurations}
\centering
\small

\begingroup
\renewcommand{\arraystretch}{1.2}

\begin{tabular}{
    >{\raggedright\arraybackslash}p{2.8cm}
    >{\raggedright\arraybackslash}p{6.0cm}
    >{\raggedright\arraybackslash}p{4.7cm}
    >{\raggedright\arraybackslash}p{2.0cm}
}
\hline
Configuration &
Training views &
Validation and test views &
Offline augmentation \\
\hline

Raw + online augmentation &
Raw image followed by label-safe letterboxing &
Raw image followed by the same letterboxing &
None \\

Offline class balancing &
Baseline view plus three generated variants for every non-\textit{nv} training image &
Identical to the raw baseline &
Train only \\

DullRazor--CLAHE &
DullRazor hair removal followed by CIELAB CLAHE and letterboxing &
The same deterministic preprocessing pipeline &
None \\

Raw--processed hybrid &
One raw view and one ROI--DullRazor--Gray World view for each training image &
The minimally processed raw view &
None \\
\hline
\end{tabular}

\endgroup
\end{table*}
DullRazor uses a $17\times17$ black-hat operator with threshold 10 \cite{lee1997dullrazor}, followed by Telea inpainting with radius 1 \cite{telea2004image}. Contrast-limited adaptive histogram equalization (CLAHE) is applied to the lightness channel of the CIE $L^{*}a^{*}b^{*}$ (CIELAB) representation using a clip limit of 2.0 and an $8\times8$ grid \cite{zuiderveld1994contrast}.

Following the Gray World assumption \cite{buchsbaum1980spatial}, let $I_c(x,y)$ denote the intensity at location $(x,y)$ in channel $c\in\{R,G,B\}$ of an image with height $H$ and width $W$. The channel mean and common target mean are given, respectively, as
\begin{equation}
\begin{aligned}
\mu_c
&=
\frac{1}{HW}
\sum_{y=1}^{H}
\sum_{x=1}^{W}
I_c(x,y),
&
\bar{\mu}
&=
\frac{1}{3}
\sum_{c\in\{R,G,B\}}
\mu_c .
\end{aligned}
\label{eq:gray_world_means}
\end{equation}

For $\mu_c>0$, color normalization is computed as
\begin{equation}
I'_c(x,y)
=
\operatorname{clip}_{[0,255]}
\left(
I_c(x,y)\frac{\bar{\mu}}{\mu_c}
\right),
\label{eq:gray_world}
\end{equation}
where $I'_c(x,y)$ denotes the normalized intensity. If any channel mean is zero, the original image is retained.

Region-of-interest (ROI) extraction uses the bounding rectangle of the largest contour. A crop is accepted only if it covers at least 50\% of the image and preserves all foreground pixels; accepted crops receive a 2\% margin.

Offline augmentation jointly transforms images and masks using flips, affine and color transformations, and Gaussian blur. Three variants are generated for each non-\textit{nv} training image using preparation seed 42; no offline variant is introduced into validation or test data.

\subsection{YOLO26 Joint Classification and Instance Segmentation}

YOLO26n-seg is initialized from \texttt{yolo26n-seg.pt} and fine-tuned for the seven lesion classes \cite{jocher2026ultralytics}. Each predicted instance contains a box, confidence, class, and mask, enabling lesion classification and instance segmentation in one inference pass.

Because each image contains one primary lesion, only the single highest-confidence instance is retained for per-image evaluation, and each ground-truth mask is reduced to its largest external contour, so both masks are single-lesion by construction. A missing detection produces an explicit \texttt{no\_detection} classification outcome and an empty predicted mask, ensuring that detection failures remain in both task evaluations.

YOLO26n-seg is fixed across the principal preprocessing strategies. YOLO11n-seg and YOLO12n-seg serve as raw-input architecture controls, while YOLO26s-seg assesses the effect of increased capacity.

\subsection{Training Protocol}

Except for the deterministic no-augmentation control, all principal configurations use the same $640\times640$ resolution, training budget, optimizer, and online augmentation policy. Training lasts up to 300 epochs with batch size 16. AdamW \cite{loshchilov2019decoupled} uses an initial learning rate of $10^{-3}$, weight decay of $5\times10^{-4}$, three warm-up epochs, and cosine decay \cite{loshchilov2017sgdr}. Early stopping uses a patience of 50 epochs, and the checkpoint with the best validation fitness is retained.

Online augmentation includes hue--saturation--value perturbation, rotations within $\pm30^{\circ}$, 10\% translation, 30\% scaling, horizontal and vertical flips, and mosaic augmentation \cite{bochkovskiy2020yolov4}; mosaic is disabled during the final ten epochs. A deterministic raw-image control disables all stochastic online augmentation.

Principal configurations are trained with seeds 42, 52, and 62 and component ablations with seed 42; validation data guide checkpoint selection, and test data are used only for final evaluation.

\subsection{Evaluation Metrics and Efficiency Measurement}

Validation and test evaluation use a confidence threshold of 0.25, an intersection over union (IoU) threshold of 0.7 for non-maximum suppression (NMS), and a mask threshold of 0.5. Let $P$ denote the highest-confidence predicted mask and $G$ the union of ground-truth polygons. The true-positive (TP), false-positive (FP), and false-negative (FN) pixel counts are
$\mathrm{TP}=|P\cap G|$,
$\mathrm{FP}=|P\setminus G|$, and
$\mathrm{FN}=|G\setminus P|$.
Following Taha and Hanbury \cite{taha2015metrics}, the pixel-level overlap metrics are given by
\begin{equation}
\mathrm{Dice}
=
\frac{2\mathrm{TP}}
{2\mathrm{TP}+\mathrm{FP}+\mathrm{FN}},
\quad
\mathrm{IoU}
=
\frac{\mathrm{TP}}
{\mathrm{TP}+\mathrm{FP}+\mathrm{FN}}.
\label{eq:pixel_metrics}
\end{equation}

Given boundary precision $P_b$, boundary recall $R_b$, and a two-pixel matching tolerance, boundary quality is measured using the boundary $F_1$ score ($\mathrm{BF}_1$), mathematically expressed as \cite{perazzi2016benchmark}
\begin{equation}
\mathrm{BF}_{1}
=
\frac{2P_bR_b}{P_b+R_b}.
\label{eq:boundary_f1}
\end{equation}

It is worth mentioning that for a missed detection, $P$ is empty, so Dice, IoU, and $\mathrm{BF}_{1}$ are zero.

Following the Microsoft Common Objects in Context (COCO) protocol \cite{lin2014microsoft}, class-aware segmentation is measured using mask mean average precision (mAP) at an IoU threshold of 0.50, denoted by $\mathrm{mAP}_{50}$, and across thresholds from 0.50 to 0.95,
denoted by $\mathrm{mAP}_{50:95}$. Ultralytics mAP evaluation uses a confidence threshold of 0.001 and an NMS IoU threshold of 0.7.

Classification metrics include accuracy, balanced accuracy, macro-F1, per-class precision, recall, F1, and confusion matrices \cite{sokolova2009systematic,brodersen2010balanced}. The \texttt{no\_detection} category is retained as an incorrect outcome but excluded from the seven-class macro average.

Efficiency is measured at batch size 1 with 20 warm-up iterations and 100 memory-resident test images; timings cover Ultralytics preprocessing, inference, and postprocessing but exclude image loading. Following \cite{bianco2018benchmark}, we report parameter count, checkpoint size, mean latency, and throughput in frames per second (FPS). For mean latency $\bar{\ell}_{\mathrm{ms}}$, FPS is computed by
\begin{equation}
\mathrm{FPS}
=
\frac{1000}{\bar{\ell}_{\mathrm{ms}}}.
\label{eq:fps}
\end{equation}
\section{Numerical Results and Discussion}

All results are obtained on the lesion-disjoint test set. Principal configurations are reported as mean $\pm$ standard deviation over the three seeds, and component ablations use seed 42.

\subsection{Main Preprocessing and Augmentation Comparison}

\begin{table*}[t]
\caption{Three-seed test performance of the main YOLO26n-seg configurations. Acc., BAcc., and BF$_1$ denote accuracy, balanced accuracy, and boundary F$_1$ score, respectively; bold values indicate the best mean.}
\label{tab:main_results}
\centering
\scriptsize

\begingroup
\renewcommand{\arraystretch}{1.22}
\setlength{\tabcolsep}{3.5pt}

\resizebox{\textwidth}{!}{%
\begin{tabular}{lcccccccc}
\hline
Input strategy &
Acc. &
BAcc. &
Macro-F1 &
Dice &
IoU &
$\mathrm{BF}_{1}$ &
Mask $\mathrm{mAP}_{50}$ &
Mask $\mathrm{mAP}_{50:95}$ \\
\hline

Raw + online augmentation &
$\mathbf{0.8210{\pm}0.0111}$ &
$\mathbf{0.6672{\pm}0.0203}$ &
$\mathbf{0.6917{\pm}0.0202}$ &
$0.9356{\pm}0.0024$ &
$0.8971{\pm}0.0027$ &
$\mathbf{0.4922{\pm}0.0048}$ &
$\mathbf{0.7364{\pm}0.0232}$ &
$\mathbf{0.5636{\pm}0.0234}$ \\

Raw without online augmentation &
$0.7753{\pm}0.0070$ &
$0.5707{\pm}0.0373$ &
$0.5893{\pm}0.0279$ &
$0.9228{\pm}0.0006$ &
$0.8815{\pm}0.0008$ &
$0.4582{\pm}0.0008$ &
$0.5654{\pm}0.0366$ &
$0.4271{\pm}0.0252$ \\

Offline class balancing &
$0.8150{\pm}0.0047$ &
$0.6654{\pm}0.0154$ &
$0.6870{\pm}0.0251$ &
$0.9332{\pm}0.0021$ &
$0.8943{\pm}0.0020$ &
$0.4841{\pm}0.0051$ &
$0.7305{\pm}0.0168$ &
$0.5614{\pm}0.0131$ \\

DullRazor--CLAHE &
$0.8026{\pm}0.0010$ &
$0.6090{\pm}0.0147$ &
$0.6356{\pm}0.0129$ &
$0.9304{\pm}0.0036$ &
$0.8900{\pm}0.0038$ &
$0.4713{\pm}0.0021$ &
$0.6940{\pm}0.0133$ &
$0.5192{\pm}0.0122$ \\

Raw--processed hybrid &
$0.8186{\pm}0.0058$ &
$0.6252{\pm}0.0161$ &
$0.6537{\pm}0.0257$ &
$\mathbf{0.9366{\pm}0.0006}$ &
$\mathbf{0.8978{\pm}0.0006}$ &
$0.4846{\pm}0.0050$ &
$0.7119{\pm}0.0089$ &
$0.5466{\pm}0.0074$ \\
\hline
\end{tabular}%
}

\endgroup
\end{table*}

\begin{figure}[t]
    \centering
    \includegraphics[width=0.96\columnwidth]
    {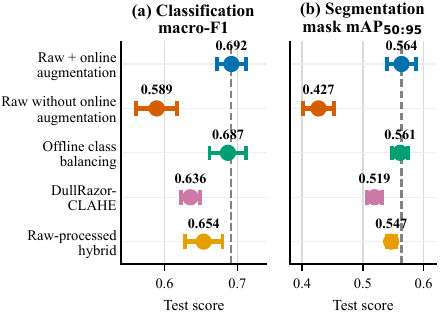}
    \caption{Main test-set comparison of the four principal input strategies and the deterministic no-augmentation control. Points denote three-seed means, horizontal error bars denote standard deviations, and dashed lines mark the raw baseline (raw + online augmentation).}
    \label{fig:main_performance}
\end{figure}

As shown in Table~\ref{tab:main_results} and summarized in Fig.~\ref{fig:main_performance}, the raw baseline (raw + online augmentation) achieves the strongest overall performance, including the highest accuracy, balanced accuracy, macro-F1, boundary F1, and mask mAP. Removing online augmentation causes the largest decline, reducing macro-F1 from $0.6917$ to $0.5893$ and mask $\mathrm{mAP}_{50:95}$ from $0.5636$ to $0.4271$, confirming that the ``less is more'' finding concerns deterministic preprocessing rather than augmentation itself. DullRazor--CLAHE reduces both classification and class-aware segmentation performance, whereas the hybrid strategy attains the best binary overlap (Dice $0.9366$, IoU $0.8978$) but lower balanced accuracy, macro-F1, and mask mAP, reinforcing that overlap quality and diagnostic recognition are not interchangeable. Given the three-seed standard deviations, offline class balancing and the hybrid stay within one standard deviation of the baseline on the class-aware metrics, so the added complexity yields no measurable joint benefit; differences among the principal configurations are interpreted descriptively rather than as significant. All configurations share the same 2.69\,M-parameter model, and the raw baseline gives the best accuracy--efficiency balance.

\subsection{Component Ablation Study}

\begin{table*}[t]
\caption{Component ablation on the lesion-disjoint test set using seed 42. The raw reference is the single-seed counterpart of the three-seed raw baseline in Table~\ref{tab:main_results}. Bold values indicate the best result in each column.}
\label{tab:ablation_results}
\centering
\scriptsize

\begingroup
\renewcommand{\arraystretch}{1.22}
\setlength{\tabcolsep}{3.7pt}

\resizebox{0.8\textwidth}{!}{%
\begin{tabular}{lcccccccc}
\hline
Configuration &
Acc. &
BAcc. &
Macro-F1 &
Dice &
IoU &
$\mathrm{BF}_{1}$ &
Mask $\mathrm{mAP}_{50}$ &
Mask $\mathrm{mAP}_{50:95}$ \\
\hline

Raw (seed 42) &
$\mathbf{0.8310}$ &
$0.6489$ &
$0.6796$ &
$0.9331$ &
$0.8943$ &
$0.4869$ &
$0.7340$ &
$0.5520$ \\

DullRazor only &
$0.8097$ &
$0.6020$ &
$0.6440$ &
$0.9329$ &
$0.8927$ &
$0.4821$ &
$0.6967$ &
$0.5224$ \\

CLAHE only &
$0.8150$ &
$0.6505$ &
$0.6623$ &
$0.9315$ &
$0.8927$ &
$0.4846$ &
$0.7366$ &
$0.5664$ \\

Gray World only &
$0.8064$ &
$0.6499$ &
$0.6730$ &
$\mathbf{0.9371}$ &
$\mathbf{0.8987}$ &
$\mathbf{0.4940}$ &
$0.7261$ &
$0.5605$ \\

ROI only &
$0.8263$ &
$\mathbf{0.6524}$ &
$\mathbf{0.6945}$ &
$0.9341$ &
$0.8962$ &
$0.4934$ &
$\mathbf{0.7405}$ &
$\mathbf{0.5729}$ \\

ROI + DullRazor + CLAHE &
$0.8097$ &
$0.6310$ &
$0.6580$ &
$0.9303$ &
$0.8902$ &
$0.4736$ &
$0.6942$ &
$0.5123$ \\

ROI + DullRazor + Gray World &
$0.8104$ &
$0.6225$ &
$0.6586$ &
$0.9293$ &
$0.8899$ &
$0.4823$ &
$0.7076$ &
$0.5396$ \\
\hline
\end{tabular}%
}

\endgroup
\end{table*}

Table~\ref{tab:ablation_results} shows that ROI extraction produces the highest balanced accuracy, macro-F1, and mask mAP, whereas Gray World achieves the best Dice, IoU, and boundary F1. CLAHE slightly improves mask mAP over the matched raw reference but reduces accuracy and macro-F1, while DullRazor consistently weakens the main metrics. Combining ROI with DullRazor and either CLAHE or Gray World produces no additive benefit and generally performs below the raw reference, indicating that successive transformations may alter useful color, texture, or boundary cues. Since these ablations use a single seed, their differences are interpreted directionally: individual operations such as ROI extraction may lead several isolated metrics, yet these gains do not survive composition and are not reproduced as three-seed principal configurations. The seed-averaged evidence in Table~\ref{tab:main_results} therefore governs the conclusion that increasingly complex preprocessing does not provide a consistent joint classification--segmentation advantage.
\subsection{Architecture and Efficiency Analysis}

\begin{table}[t]
\caption{Raw-input architecture and efficiency comparison. M and MiB denote millions of parameters and mebibytes; results are mean $\pm$ standard deviation over three seeds, and YOLO26s-seg is reported for seed 42. FPS is the per-seed mean of \eqref{eq:fps}, which differs slightly from $1000/\bar{\ell}_{\mathrm{ms}}$ as the reciprocal and mean do not commute.}
\label{tab:architecture_efficiency}
\centering
\scriptsize

\begingroup
\renewcommand{\arraystretch}{1.22}
\setlength{\tabcolsep}{3.5pt}

\resizebox{\columnwidth}{!}{%
\begin{tabular}{lcccccc}
\hline
Architecture &
Params (M) &
Size (MiB) &
Macro-F1 &
Mask $\mathrm{mAP}_{50:95}$ &
Latency (ms) &
FPS \\
\hline

YOLO11n-seg &
2.84 &
5.76 &
$0.6734{\pm}0.0068$ &
$0.5501{\pm}0.0150$ &
$25.51{\pm}4.39$ &
$39.92{\pm}6.28$ \\

YOLO12n-seg &
2.78 &
5.74 &
$0.6962{\pm}0.0178$ &
$0.5758{\pm}0.0067$ &
$20.17{\pm}6.65$ &
$54.33{\pm}21.76$ \\

YOLO26n-seg &
2.69 &
6.28 &
$0.6917{\pm}0.0202$ &
$0.5636{\pm}0.0234$ &
$20.48{\pm}4.24$ &
$50.14{\pm}9.42$ \\

YOLO26s-seg &
10.37 &
22.29 &
$0.7239$ &
$0.5945$ &
$14.26$ &
$70.13$ \\
\hline
\end{tabular}%
}

\endgroup
\end{table}

Among the nano-scale models, YOLO12n-seg achieves the highest mean macro-F1 ($0.6962$) and mask $\mathrm{mAP}_{50:95}$ ($0.5758$), whereas YOLO26n-seg uses the fewest parameters (2.69 M). YOLO12n-seg attains a measured throughput of $54.33\pm21.76$ FPS, but the large run-to-run variation, which reflects system contention on a single shared GPU, together with the separately executed benchmarks, makes latency rankings descriptive. YOLO26s-seg further improves the seed-42 scores but requires 3.85 times as many parameters and a checkpoint 3.55 times as large as YOLO26n-seg. Overall, YOLO12n-seg and YOLO26n-seg are statistically indistinguishable on the accuracy metrics across three seeds (overlapping standard deviations), with YOLO26n-seg the most parameter-efficient; this equivalence does not alter the preprocessing comparison, for which YOLO26n-seg is fixed across all strategies.

\subsection{Per-Class and Qualitative Analysis}

\begin{figure}[t]
    \centering
    \includegraphics[width=0.80\columnwidth]
    {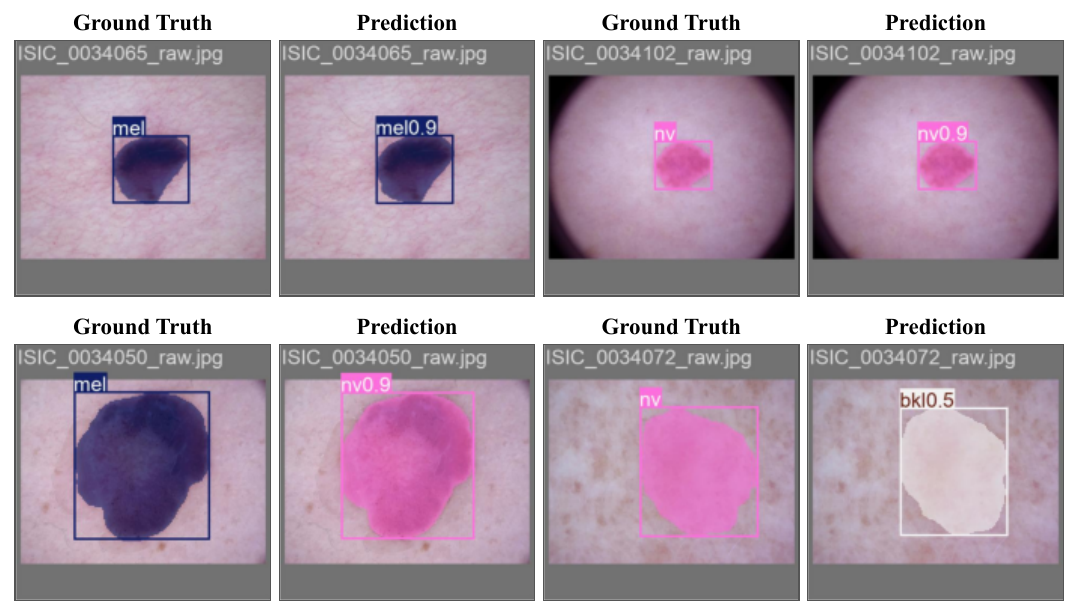}
    \caption{Qualitative test predictions of the raw YOLO26n-seg baseline using seed 52. The upper examples are correctly classified, whereas the lower examples retain high mask overlap but receive incorrect diagnostic labels.}
    \label{fig:qualitative_predictions}
\end{figure}

For the raw baseline, the mean per-class F1 scores over three seeds are $0.9127$ for \textit{nv}, $0.8008$ for \textit{vasc}, $0.7449$ for \textit{bcc}, $0.7238$ for \textit{df}, $0.6971$ for \textit{bkl}, $0.5423$ for \textit{mel}, and $0.4204$ for \textit{akiec}. The low \textit{akiec} F1 is primarily associated with a recall of $0.2789$ despite a precision of $0.8585$, meaning that most \textit{akiec} lesions are missed and assigned to other classes; this is a safety-relevant limitation for a pre-malignant category, driven by its low support and overlap with \textit{bkl}. The confusion matrices further show that the dominant melanoma error is \textit{mel}$\rightarrow$\textit{nv} (melanoma predicted as benign nevus), the most safety-critical failure mode, together with \textit{akiec}$\rightarrow$\textit{bkl} confusion, consistent with class imbalance and overlapping dermoscopic appearance.
Fig.~\ref{fig:qualitative_predictions} illustrates this task dependence: the model accurately delineates representative \textit{mel} and \textit{nv} lesions, but some highly overlapping masks are assigned to the wrong diagnostic class. In particular, Dice scores of $0.991$ and $0.936$ accompany \textit{mel}$\rightarrow$\textit{nv} and \textit{nv}$\rightarrow$\textit{bkl} errors, respectively, confirming that geometric agreement alone is insufficient for evaluating joint lesion analysis and that a segmentation-only score would conceal these diagnostic errors.

\subsection{Discussion and Limitations}

The results favor minimally processed dermoscopic images combined with online augmentation: only online augmentation yields a broad gain, whereas deterministic preprocessing trades isolated overlap improvements for weaker class-aware recognition. Thus, ``less is more'' refers to limiting deterministic preprocessing rather than removing augmentation. These results characterize a research baseline for method comparison and do not establish diagnostic or triage utility; the high Dice partly reflects the single-lesion, lesion-filling geometry of HAM10000 rather than clinically difficult boundary cases. The findings also remain bounded by the HAM10000 acquisition domain, class imbalance, and limited minority-class support. Moreover, three seeds provide limited evidence for small differences, while the ablations and YOLO26s control are directional seed-42 comparisons. The single-lesion prediction policy reflects the dataset structure but not multi-lesion scenes, and the reported throughput excludes image loading and offline preprocessing, representing model execution rather than complete workflow latency.

\section{Concluding Remarks and Open Problems}

This study presented a leakage-controlled, lesion-disjoint evaluation of dermoscopic preprocessing for joint multi-class classification and instance segmentation. Under a fixed YOLO26n-seg with all non-input factors held constant, minimally processed images with online augmentation delivered the strongest overall balance of classification, class-aware segmentation, boundary quality, and efficiency (2.69\,M parameters, approximately 50 FPS), whereas increasingly complex deterministic preprocessing produced no consistent joint benefit across three seeds. This does not imply that preprocessing is unhelpful in general: individual operations improved isolated metrics, and cross-device acquisition may shift the balance. The architecture controls showed YOLO12n-seg and YOLO26n-seg to be statistically indistinguishable on accuracy, with YOLO26n-seg the most parameter-efficient. These results establish a reproducible lightweight baseline and emphasize evaluating recognition and segmentation together. Future work should examine cross-dataset and cross-device generalization, improve minority-class recognition, extend evaluation to multi-lesion settings, and investigate adaptive or learnable normalization under the same leakage-controlled protocol.

\section*{Acknowledgment}

This research is funded by Van Lang University, Vietnam under grant number 2510-DT-VLT-KCT-SV-003.
\bibliographystyle{IEEEtran}
\bibliography{IEEEabrv,reference}

\end{document}